%% file: main.tex
\documentclass[10pt,twocolumn,letterpaper]{article}

\usepackage[final,applications]{wacv}

\usepackage{times}
\usepackage{epsfig}
\usepackage{graphicx}
\usepackage{amsmath}
\usepackage{amssymb}
\usepackage{booktabs}
\usepackage{xcolor}
\usepackage{multirow}

\def\confName{WACV}
\def\confYear{2027}

\usepackage[pagebackref,breaklinks,colorlinks,citecolor=blue,bookmarks=false]{hyperref}
\usepackage{microtype}

\begin{document}

\title{Reliability-Gated Fusion of Consumer Head and Foot IMUs\\for Lower-Body 3D Pose}

\author{Zhilin Guo$^{1*}$ \quad Boqiao Zhang$^{1*}$ \quad Oszk\'ar Urb\'an$^{1*}$ \quad Josef Bengtson$^{2*}$ \quad Hakan Aktas$^{1}$ \quad Wenzhao Li$^{1}$\\
Siyu Hong$^{1}$ \quad Kyle Fogarty$^{1}$ \quad Chenliang Zhou$^{1}$ \quad Ali Senguel$^{1}$ \quad Cengiz Oztireli$^{1}$\\[4pt]
$^{1}$University of Cambridge \qquad $^{2}$Chalmers University of Technology\\
{\tt\small aco41@cam.ac.uk}\\[2pt]
{\normalsize $^{*}$Equal contribution}
}

\maketitle

\begin{figure*}[t]
\centering
\includegraphics[width=\textwidth]{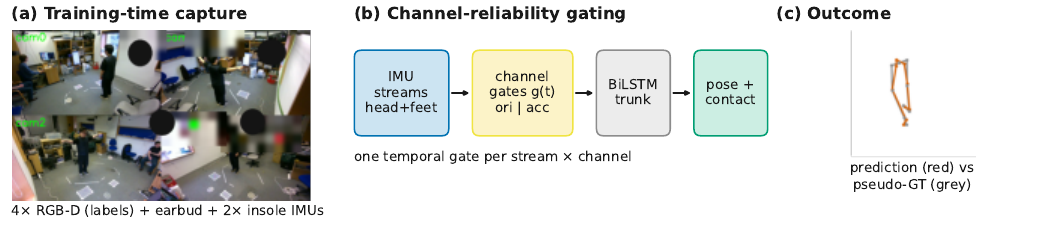}
\caption{Overview. (a) Training-time capture: four RGB-D views (labels only), an earbud head IMU, and two smart-insole foot IMUs, with no pelvis sensor and no foot gyroscope. (b) Channel-reliability gating: one temporal gate per stream per channel block learns how much, and when, to trust orientation vs.\ acceleration, supervised only by synthetic corruption in pretraining. (c) Held-out prediction (red) vs.\ pseudo-GT (grey): foot acceleration alone reaches 66.6\,mm (79.0\,mm head-only), the channel gate is the most accurate of our learned fusion arms in every fault regime, and the gates flag dropout bursts at 0.92--0.999 AUROC without test-time supervision.}
\label{fig:teaser}
\end{figure*}

\begin{abstract}
Sparse inertial pose estimation promises camera-free motion capture from consumer devices, but consumer sensors are unreliable: firmware-fused orientations are biased, mounting varies between sessions, and streams drift or drop out. On a new 35-take single-subject benchmark pairing an earbud head inertial measurement unit (IMU) with two smart-insole foot IMUs (SAM-3D-Body pseudo-ground-truth labels), we show the reliability problem is channel-level: a channel ablation isolates foot acceleration as the most informative input (66.6\,mm vs.\ 79.0\,mm head-only) and the firmware-fused foot orientation as the liability that destroys the gain. We therefore let the model learn how much to trust each channel of each stream: one temporal gate per stream per channel block, trained with an auxiliary reliability objective on synthetically corrupted pretraining data. The channel-gated model is the most accurate of our learned fusion arms on clean data (69.4\,mm vs.\ 83.7 static, 86.6 ungated) and under every simulated fault (bias in training; drift, dropout eval-only); its gates suppress the natively biased foot-orientation channels on clean real data without test-time supervision and flag dropout bursts at 0.92--0.999 AUROC. Two contrasts: dropping a channel known a priori to fail is flat across foot faults but collapses when an unanticipated stream fails (head dropout: 92.9 vs.\ 79.3\,mm); and a fine-tuned HMD-Poser is more accurate on clean data (64.4\,mm) and nominally under drift, with no significant paired difference under bias or dropout, but a larger worst-case degradation from clean ($+16.1$ vs.\ $+3.5$\,mm, single seed). Learning to gate reliability instead of sensor count is the lever for deployable sparse inertial capture. Code is available at \url{https://github.com/ZhilinGuo/reliability-gated-imu-fusion}.
\end{abstract}

\section{Introduction}
\looseness=-1
Sparse wearable pose estimation promises camera-free avatar control, motion entertainment, and at-home gait analysis: once trained, an inertial model needs no cameras at deployment. Yet most published systems assume a six-IMU suit, a pelvis or back tracker, and research-grade sensors with raw gyroscopes~\cite{huang2018dip,yi2021transpose,yi2022pip,jiang2022tip}. We instead target a configuration built entirely from consumer devices (Fig.~\ref{fig:teaser}): one head IMU from earbuds (AirPods) and two foot IMUs from smart insoles (Striv). The layout suits gait and lower-body motion: feet observe stance, swing, and contact; the head provides a stable orientation reference. But it breaks two assumptions the literature relies on: there is no pelvis/root sensor, so root translation and hip axial rotation are weakly observed; and the insole firmware emits a fused Euler orientation with no raw gyroscope, so calibration error cannot be removed by gyro-based filtering.

With neither marker-based mocap nor hardware synchronization, we treat label generation and alignment as first-class system problems. Cameras are required at training time only: vision is the teacher, not the input; the one deployment-time vision dependency is per-take mounting calibration, isolated and bounded in Sec.~\ref{sec:external}. We capture four-view RGB-D video alongside the wearables, align modalities after the fact by motion agreement~\cite{li2014temporal}, and generate pseudo-ground-truth (pseudo-GT) with SAM 3D Body~\cite{sam3dbody}. We then benchmark an IMUPoser-style recurrent model~\cite{imuposer} pretrained on synthetic AMASS IMU~\cite{mahmood2019amass} and fine-tuned on our pseudo-GT.

\looseness=-1
This paper extends a preliminary workshop pilot~\cite{bkpilot} on the same takes, which reported that adding the foot IMUs degrades pose below a head-only model. On the corrected pipeline the penalty shrinks below detectability on held-out runs (80.3 vs.\ 79.0\,mm sim-MPJPE; paired per-take median $+1.8$\,mm, $p{=}0.41$). A channel ablation reveals a sharp internal structure: foot acceleration alone is the best input configuration we test (66.6\,mm, significantly better than head-only), foot orientation alone is worse than no feet on average (marginally so two-sided, $p{=}0.054$), and adding the orientation channels to the acceleration channels destroys most of the gain (Hodges--Lehmann $+12.9$\,mm, $p{<}10^{-5}$). The prize and the toxin arrive in the same stream. Since deployment faults are unknown a priori and time-varying, the research question shifts from ``how many sensors?'' to ``how much should each channel of each sensor be trusted, and when?''

\looseness=-1
Our answer is \emph{channel-reliability gating}: each channel block (orientation, acceleration) of each stream is weighted by a scalar gate with a short temporal receptive field, trained with an auxiliary reliability objective during corruption-augmented pretraining. The channel gate is the best learned fusion arm on clean data and under all four simulated fault regimes, and its gates are interpretable without test-time supervision: on clean real data they partially close on the natively biased foot-orientation channels, they close further when bias enters the fine-tuning distribution, and they react online to burst faults. A control trained without the auxiliary objective leaves gates high and unselective on the same clean data: the selectivity is transferred from synthetic-corruption supervision, not emergent from the task loss. Stream-level gates add little over static learned weights, an honest negative result that localizes the benefit at channel granularity.

\noindent\textbf{Contributions.}
\begin{itemize}\itemsep2pt
\item A 35-take single-subject consumer-IMU benchmark (4$\times$RGB-D + earbud head IMU + two insole IMUs) with released capture, after-the-fact alignment, and pseudo-GT labeling protocols, and run/motion held-out evaluation with similarity-aligned, distal, and temporal metrics.
\item A controlled channel-level diagnosis of consumer foot IMUs: foot acceleration alone is the best input configuration (66.6\,mm), foot orientation alone is toxic, and a monotone dose--response isolates orientation bias, not sensor count, as the cause.
\item Channel-reliability gating: one temporal gate per stream per channel block with corruption-supervised pretraining, the most accurate learned-fusion arm on clean data and under all fault regimes (69.4\,/\,77.8\,/\,69.6\,/\,72.8\,mm clean/bias/drift/dropout, three seeds; sign-consistent in $7/7$ classes); gates suppress the natively biased channels without test-time supervision and flag dropout bursts at 0.92--0.999 AUROC.
\item A deployment analysis bounding the design space: a priori channel drop vs.\ learned gating under anticipated and unanticipated faults, an external HMD-Poser comparison at $4.6\times$ CPU throughput with $37\%$ fewer parameters, and negative results for one-time T-pose calibration, oracle hard masking, and augmentation without gating.
\end{itemize}

\section{Related Work}
\noindent\textbf{Sparse-IMU full-body pose.}
Sparse-IMU reconstruction progressed from offline optimization over a statistical body model~\cite{vonmarcard2017sip} through learning-based six-IMU pipelines~\cite{huang2018dip}, real-time physics-aware optimization~\cite{yi2021transpose,yi2022pip}, Transformers~\cite{jiang2022tip}, diverse-real-data generalization~\cite{zhang2024dynaip}, and diffusion from arbitrary sensor sets including pressure insoles~\cite{vanwouwe2024diffusionposer}. These define the methodological backbone but assume a pelvis sensor and raw gyroscopes.

\noindent\textbf{Commodity-device and VR sparse tracking.}
IMUPoser, our primary baseline for its transparency and earbud relevance, estimates pose from variable phone/watch/earbud subsets~\cite{imuposer}; MobilePoser targets 1--3 mobile IMUs~\cite{mobileposer}, ProgIP uses three IMUs at head and wrists~\cite{progip}, and IMUCoCo conditions on continuous on-body coordinates~\cite{imucoco}. VR trackers infer full-body motion from head-and-hands input~\cite{jiang2022avatarposer,du2023agrol,jiang2024egoposer,yang2024divatrack}; HMD-Poser scales to HMD plus a few IMUs with pretrained HMD+2IMU weights~\cite{hmdposer}. These observe the upper body and hallucinate the legs, the inverse of our feet-informed setting; none places sensors at head plus both feet.

\noindent\textbf{Foot/insole sensing and clinical gait.}
Reduced-IMU lower-limb estimation recovers walking kinematics from two or three IMUs at ${\sim}6$\,cm segment error while degrading on turns and dynamic motion~\cite{sy2021liegroup,hori2025gip}; GRIP fuses insole pressure with a physics digital twin~\cite{grip2026}; wearables see growing use in Parkinsonian gait assessment~\cite{fog2025review}. This literature both motivates feet and predicts where they fail.

\noindent\textbf{Robustness and gating.}
Sensor faults are an emerging concern: TIC learns dynamic on-body calibration~\cite{tic2025}, MagShield hardens against magnetic disturbance~\cite{magshield2025}, UMotion propagates inertial+ultra-wideband (UWB) uncertainty~\cite{umotion2025}, and WHIP, the closest concurrent work, reconstructs motion from consumer wearables including instrumented insoles, but with raw IMU channels and hardware-synchronized mocap supervision~\cite{whip2026}. Our setting is complementary and harsher: consumer fused orientations whose bias is not identifiable under the training distribution, faults that appear only at test time, and a gate that must infer reliability from the stream itself. Gating is classical for modality fusion~\cite{arevalo2017gmu,shazeer2017moe,han2021trusted,hu2018se,perez2018film,srivastava2015highway}; we claim no new mechanism; the contribution is its instantiation and validation for test-time sensor faults in inertial capture, where the reliability signal must come from the faulty stream itself.

\noindent\textbf{Vision+IMU fusion and pseudo-GT.}
DeepFuse, the closest WACV precedent, fuses body-worn IMUs with multi-view images~\cite{huang2020deepfuse}; pose-based multi-view alignment removes hardware sync~\cite{delattre2026alignment}. We use vision only to create labels: SAM 3D Body for pseudo-GT~\cite{sam3dbody,fastsam3dbody}, AMASS for synthetic pretraining~\cite{mahmood2019amass}, motion agreement for alignment~\cite{li2014temporal}. To our knowledge no prior work benchmarks our layout: head + two feet, no pelvis, no foot gyro.

\section{Capture, Alignment, and Pseudo-Ground-Truth}
\label{sec:data}

The capture rig comprises four Intel RealSense RGB-D cameras ($848\times480$ at 30\,FPS, color + depth), one AirPods head IMU streamed via a custom iOS app (fused attitude, device-frame linear acceleration, gyroscope), and left/right Striv insole IMUs (fused Euler orientation plus acceleration, no raw gyroscope, ${\sim}30$\,Hz over BLE). The motion protocol is five repeated runs of seven scripts (35 takes; supp.~S1) spanning standing and straight walking with $180^\circ$ turns, in-place stepping and single-leg stance, side-steps/grapevine/$360^\circ$ turns, squats/bends/lunges/sit-to-stand, an everyday walk--sit--pick-up composite, clinical gait (abrupt stop/restart, dual-task), and upper-limb reaches and strikes. The protocol is intentionally dual-use, covering clinical gait and turning as well as avatar-oriented composite motion.

\noindent\textbf{Dataset statistics.}
The benchmark contains 82{,}417 frames (45.8 minutes at 30\,FPS) across the 35 takes (mean 78.5\,s, range 38.6--137.1\,s). SAM 3D Body yields a usable nine-joint lower-body label for every frame; each foot is in ground contact for 65\% of frames (73\% for at least one foot). Each take stores aligned head and foot orientation matrices and accelerations (and head gyro, used only for alignment and in one ablation arm), the nine lower-body joints, a 70-joint Momentum-Human-Rig pose, per-foot contacts, a global root rotation, and a validity mask. \textbf{Human-subjects statement:} the single subject is an author who self-captured with informed consent; no identifiable imagery is published.

\noindent\textbf{Temporal alignment.}
We have no hardware sync, only Unix-style timestamps with constant offsets and slow drift. We resample all streams onto a common 30\,Hz grid, then estimate two offsets. The \emph{head} offset cross-correlates AirPods gyro magnitude against head angular speed derived from SAM orientation, following motion-agreement temporal calibration~\cite{li2014temporal}. A \emph{shared foot} offset (the insoles share one BLE base station and clock) cross-correlates Striv motion intensity against SAM foot speed over the whole take. The full-dataset offsets ($-2.69{\pm}0.06$\,s head, $-3.05{\pm}0.10$\,s feet) are consistent with the expected clock skew, and excluding the six takes whose head-alignment confidence falls below $0.5$ leaves every headline conclusion unchanged (supp.~S1).

\noindent\textbf{Pseudo-ground-truth via SAM 3D Body.}
For supervision we run SAM 3D Body~\cite{sam3dbody} on a selected camera view and convert its Momentum-Human-Rig output to a nine-joint lower-body target: pelvis, left/right hip, knee, ankle, and foot (supp.~S2). We treat these labels as pseudo-GT with measurable bias rather than mocap-grade truth: single-image recovery is strong on clear standing and walking frames but noisier under self-occlusion, fast motion, and turning. As a label-integrity control, we re-labeled all 35 takes with four-view fused labels (per-view residuals 7.9--27.9\,mm, median 15.0) and re-scored the same held-out head-only predictions against both label sets, with no retraining: lower-body error changes by $0.6$\,mm (83.9 vs.\ 84.5\,mm; supp.~S2), so label view choice does not drive any conclusion. Both label sets derive from the same model family, so this control validates view choice, not absolute label correctness (Sec.~\ref{sec:discussion}). Each take is spatially calibrated by fitting a constant sensor-to-bone rotation per sensor against that take's own SAM-derived bone frames over the whole take (a transductive fit; two-sided Procrustes; fitted magnitudes and audits: supp.~S1).

\section{Method}
\label{sec:method}
\begin{figure}[t]
\centering
\includegraphics[width=\linewidth]{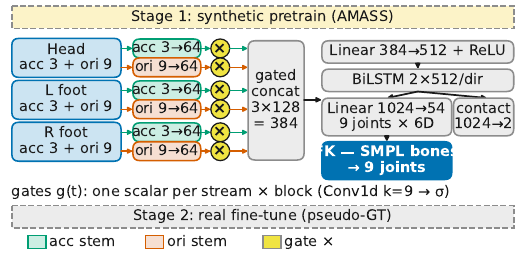}
\caption{Reliability-gated fusion. Per-stream stems (head: earbud; feet: smart insoles; split per channel block) feed scalar temporal gates $g_{i,b}(t)$, a shared Conv1d ($k{=}9$) $\to$ ReLU $\to$ sigmoid producing one scalar per stream$\times$block. Gated features are concatenated and projected into a two-layer bidirectional LSTM trunk ($2{\times}512$/direction) regressing 6D rotations for nine lower-body joints, then forward kinematics with fixed SMPL bone lengths and a contact head. Stage 1: pretrain on synthetic AMASS with foot corruption ($p{=}0.5$) and a gate BCE objective. Stage 2: fine-tune on real pseudo-GT with an aligned-FK $\ell_1$ loss (gate loss off).}
\label{fig:arch}
\end{figure}
Our model augments an IMUPoser-style recurrent backbone~\cite{imuposer} with a reliability-gated fusion layer (Fig.~\ref{fig:arch}). Each IMU contributes linear acceleration and a $3\times3$ orientation matrix, packed as a 12-D per-sensor vector; the full input is 36-D for head-plus-feet, and sensor ablations zero the unused blocks. A two-layer bidirectional LSTM maps the windowed input to 6D rotations~\cite{zhou2019continuity} for nine lower-body joints, and forward kinematics with SMPL-style conventions~\cite{loper2015smpl} produces 3D joint positions. We pretrain on synthetic IMU generated from CMU, BioMotionLab NTroje, and MPI HDM05 sequences in AMASS~\cite{mahmood2019amass} by placing virtual sensors at the head and both feet (omitting any synthesized foot gyro to match Striv), then fine-tune per fold on real pseudo-GT.

\noindent\textbf{Reliability-gated fusion.}
Fixed concatenation forces the trunk to trust every stream equally; we instead let the model down-weight an untrustworthy channel. Each stream $i$ (head, left foot, right foot) is mapped by a linear stem to a 128-D feature $h_i(t)$; for the channel-level variant the stem is split per channel block, acceleration ($3{\to}64$) and orientation ($9{\to}64$), so that $h_i=[h_{i,\mathrm{acc}}\|h_{i,\mathrm{ori}}]$ keeps the two blocks separable. Gates $g_{i,b}(t)\in(0,1)$ estimate the instantaneous reliability of channel block $b$ (orientation or acceleration) of stream $i$, each applied to its own block's features, and the gated features are concatenated and projected:
\begin{equation}
z(t)=W_c\,\big[\,g_{1}(t)\,h_{1}(t)\;\|\;g_{2}(t)\,h_{2}(t)\;\|\;g_{3}(t)\,h_{3}(t)\,\big],
\end{equation}
omitting the block index $b$ where the gate is per-stream. Concatenation (rather than a gated sum) lets the trunk observe each stream's gated feature separately; a gated-sum variant is ablated in Sec.~\ref{sec:gated}. Gates depend only on each stream's own stem features; the frozen, static, and per-frame gates are causal by construction, and the temporal gates become causal with causal padding (Sec.~\ref{sec:external} evaluates a fully causal variant). A dropped stream presents as zeroed input frames, which the dropout corruption below trains the gates to close on.

We compare five gate instantiations. \emph{Frozen} ($g{=}1$) is an architecture-matched no-gate control. \emph{Static} learns one scalar per stream. \emph{Per-frame learned} predicts $g_i(t)=\sigma(w^\top h_i(t))$ from the current frame. \emph{Temporal} adds a short temporal receptive field, $g_i(t)=\sigma(w^\top\mathrm{ReLU}(\mathrm{Conv1d}_9(h_i))(t))$ with kernel $9$ and symmetric padding ($\pm133$\,ms, non-causal). \emph{Channel-level temporal}, our variant, runs the same temporal gate over each stream's concatenated block features and emits one scalar gate per block, $g_{i,b}(t)$, each scaling its own block's 64-D half of $h_i$, so the model can keep a stream's acceleration while shutting off its orientation: a constant bias or slow drift, invisible in any single frame, is visible in orientation statistics over a short window.

\noindent\textbf{Corruption augmentation and gate supervision.}
Pretraining synthesizes three corruption types on random foot streams with per-window probability $0.5$: constant \emph{bias} (a fixed body-frame rotation offset per window), \emph{drift} (a slowly varying offset), and \emph{dropout} (bursts of zeroed frames), each accompanied by an indicator $c_i(t)$. Gates receive an auxiliary binary cross-entropy (BCE) objective (weight $0.1$) against per-block reliability $1-c_{i,b}(t)$, where bias and drift mark only the orientation block corrupted and dropout marks both. This supervision applies during pretraining only: on real fine-tuning data the ``clean'' foot streams are natively biased, and we found that continuing gate supervision there forces gates open against the task loss. Fine-tuning therefore optimizes the aligned-FK position loss alone, with the same corruption augmentation kept active so the gates retain their pretrained selectivity (Sec.~\ref{sec:interp} verifies that this selectivity is transferred from the auxiliary supervision, not emergent from the task loss).

\noindent\textbf{Implementation details.}
\looseness=-1
The network is a faithful IMUPoser \texttt{RNN}: per-stream stems, the gated fusion layer, a linear projection to 512 units, a two-layer bidirectional LSTM (512 hidden units per direction, input dropout 0.2), and a linear head to the $9\times6=54$-D pose output. Acceleration is globally scaled by $30\,\mathrm{m/s^2}$, matching IMUPoser. Synthetic pretraining uses an $\ell_1$ loss on the 6D rotations, Adam (lr $3{\times}10^{-4}$), 40 epochs, reaching $6.9^\circ$ mean geodesic error on held-out synthetic motion. Fine-tuning uses 5\,s windows (150 frames), batch 32, Adam (lr $1{\times}10^{-4}$), 20 epochs, evaluating the final checkpoint; the loss is computed in 3D after differentiable forward kinematics under a detached per-window similarity (rotation + scale) alignment, rewarding correct relative lower-body geometry rather than absolute placement.

\section{Experimental Setup}
\label{sec:setup}
We evaluate two protocols. \emph{Leave-one-run-out} (``run'') holds out one of the five repeats and tests on unseen repetitions of seen motions; \emph{leave-one-motion-out} (``seq'') holds out an entire motion class and is the harder generalization test. All numbers use the full 35-take checkpoint sweep, and we report a constant mean-pose prior and a zero-input control as sanity floors.

\noindent\textbf{Metrics.}
The model predicts joint rotations only; positions come from forward kinematics with fixed bone lengths, and the pseudo-GT is stored root-relative, so root translation is neither predicted nor evaluated and all position metrics are root-relative. Let $\hat{p}_{t,j},p_{t,j}\in\mathbb{R}^3$ be predicted and pseudo-GT positions of joint $j$ at frame $t$. Per-frame Procrustes PA-MPJPE is too forgiving here, since a static mean pose scores well after per-frame alignment, so we emphasize \emph{sim-MPJPE}, which fits a single per-sequence scale and rotation $(s,R)$ on the root-relative positions (no translation degree of freedom):
\begin{equation}
\mathrm{sim\text{-}MPJPE}=\min_{s,R}\frac{1}{TJ}\sum_{t,j}\big\| sR\,\hat{p}_{t,j}-p_{t,j}\big\|_2 .
\end{equation}
The scale degree of freedom is near-redundant for our models (fixed FK bone lengths) and exists to accommodate external baselines whose output scale is miscalibrated (Sec.~\ref{sec:external}). We also report per-joint \emph{motion} $R^2_j = 1-\sum_t \|\hat{p}_{t,j}-p_{t,j}\|_2^2 / \sum_t \|p_{t,j}-\bar{p}_{j}\|_2^2$ ($\bar{p}_j$ the temporal mean, computed on the sim-aligned predictions; scalar summaries average the nine joints), i.e., the fraction of joint-motion variance explained, which penalizes a static prediction. \emph{Distal} variants average only knees, ankles, and feet, where the foot sensors should matter most. MPJVE measures velocity error from finite differences, and foot-contact quality uses precision/recall/F1 against SAM-derived contacts.

\noindent\textbf{Corruption regimes.}
\looseness=-1
We evaluate three synthetic foot-stream fault types, mirroring the pretraining corruptions of Sec.~\ref{sec:method}: \emph{bias}, a constant per-take SO(3) mounting offset applied in training and testing, drawn per axis from $\mathcal{N}(0,\sigma)$ with $\sigma\in\{10^\circ,20^\circ\}$ (mean magnitude ${\approx}1.6\sigma$); \emph{drift}, a slowly time-varying orientation offset (random walk, $0.1^\circ$/frame, capped at $30^\circ$), test-time only; and \emph{dropout}, foot packets zeroed in random 0.5--2\,s bursts covering ${\sim}10\%$ of frames (realized 10.8\%), test-time only. Eval-only bias probes (head-fault, dose--response, $\sigma{=}30^\circ$ external comparison) instead draw the magnitude ${\sim}U(0,\sigma)$ about a uniform axis; the two operators share only the nominal $\sigma$ and are never compared without conversion (supp.~S3). All gated-fusion numbers are three-seed means on the run split. For significance, seeds share data and the pretrained initialization and differ only in fine-tuning stochasticity (batch order, dropout, augmentation draws), so we average each take's error across seeds and run paired per-take Wilcoxon signed-rank tests ($n{=}35$), with Holm correction across the declared family of 15 arm$\times$regime contrasts against the ungated model (fixed before analysis, predating the attention arm, which is therefore excluded; full table: supp.~S4) and all other contrasts reported as exploratory. Because takes are clustered by motion class, we additionally report script-level sign-consistency tests ($n{=}7$) as the cluster-robust evidence level; take-level $p$-values are anti-conservative under clustering and should be read with the reported effect sizes (Hodges--Lehmann 95\% CIs for headline contrasts, medians elsewhere; $\pm$ is population seed s.d.).

\section{Results}
\label{sec:results}
\subsection{The problem is channel-level}
\label{sec:problem}
\begin{figure}[t]
\centering
\includegraphics[width=\linewidth]{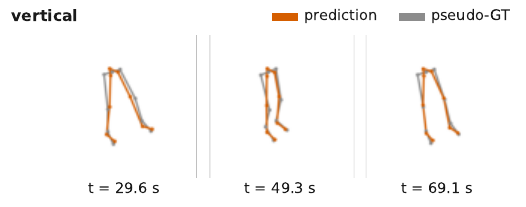}
\caption{Channel-gated predictions (red) vs.\ SAM-3D-Body pseudo-GT (grey) on a held-out vertical-motion take (squats and lunges), three frames. The full three-class montage and a head-only contrast are in supp.~S4.}
\label{fig:qual}
\end{figure}
\begin{table*}[t]
\centering
\caption{Full 35-take benchmark. Position errors (sim-MPJPE, distal, PA-MPJPE) in mm; distal averages knees, ankles, and feet; MPJVE in mm/s; ``run''/``seq'': leave-one-run-out / leave-one-motion-out. Full foot streams cost nothing detectable; removing mounting calibration degrades accuracy badly.}
\label{tab:main}
\resizebox{\textwidth}{!}{\input{tables/main_results.tex}}
\end{table*}

\looseness=-1
In Table~\ref{tab:main}, head-only input reaches 79.0\,mm sim-MPJPE on the run split and 92.6\,mm on the harder seq split. Adding the full foot streams produces no detectable penalty on either split (80.3 / 97.6\,mm; paired per-take median $+1.8$\,mm run, $p{=}0.41$; $+1.8$\,mm seq, $p{=}0.24$): the large penalty reported in the pilot does not survive a corrected stream parser (supp.~S1). This is a failure to detect a difference, not proof of equivalence, though the interval bounds any penalty well below the label-noise floor (Hodges--Lehmann $+2.5$\,mm, 95\%\,CI $[-3.0,+7.4]$ run; $+2.8$\,mm $[-1.7,+8.6]$ seq); a $\pm5$\,mm equivalence test (margin at the label-noise floor) is not significant (two one-sided tests, TOST $p{=}0.16$), and on the seq split harm concentrates in the stepping, turning, and composite classes (supp.~S4). Controls bound the effect: zero-input (116.4\,mm) and mean-pose (105.5\,mm) floors trail head-only, so the network uses real inertial signal; zero-shot transfer is poor (316.5\,mm); without AMASS pretraining, 95.3\,mm; a MobilePoser-adapted second family reproduces the parity ordering (85.5 vs.\ 86.9\,mm; full table: supp.~S4). Removing the mounting calibration (raw device streams; supp.~S1), in contrast, is unambiguous: error jumps to 103.2\,mm ($+22.8$ over calibrated head+feet), and the native mounting offsets are large: median $34^\circ$ fitted sensor-to-bone rotation for the feet (range $9^\circ$--$46^\circ$), tilt-dominated and confirmed by gravity/depth checks (supp.~S1).

\noindent\textbf{The cause is the orientation channel, not the sensor.}
\looseness=-1
A channel ablation separates the informative signal from the toxic one (three seeds each; table: supp.~S4). On the head stream, orientation alone is insufficient (93.9\,mm), orientation+acceleration recovers the head-only result (78.5\,mm; cf.\ 79.0\,mm in Table~\ref{tab:main}), and adding the head gyroscope changes nothing (78.7\,mm). On the foot streams the split is decisive: foot acceleration only yields the best configuration tested (66.6\,mm, HL $-10.7$\,mm vs.\ the Table~\ref{tab:main} head-only model, 95\%\,CI $[-16.0,-6.7]$, $p{=}1.2{\times}10^{-5}$), foot orientation only is worse than no feet on average (83.9\,mm; HL $+4.8$\,mm $[-0.0,+9.3]$), though only marginally so two-sided ($p{=}0.054$), and adding orientation to acceleration destroys most of the gain (66.6$\rightarrow$80.3\,mm; HL $+12.9$\,mm $[+8.8,+19.3]$, $p{=}8.7{\times}10^{-6}$). The prize is foot acceleration, specifically the high-passed world-frame signal (whose world rotation still consumes the uncorrected orientation, supp.~S1); the toxin is the firmware-fused foot orientation. A dose--response sweep of injected constant mounting bias degrades the ungated model monotonically on every seed (86.6 to 114.7\,mm from $\sigma{=}0^\circ$ to $40^\circ$, Spearman $\rho{=}1.0$; supp.~S4)~\cite{sy2021liegroup,hori2025gip}. When the toxic channel is known a priori, dropping it is robust by construction (orientation faults never reach the model); Sec.~\ref{sec:external} quantifies what that costs when the other stream fails.

\subsection{Reliability-gated fusion}
\label{sec:gated}
\begin{table}[t]
\centering
\caption{Gated-fusion matrix (sim-MPJPE mm, run split, three-seed mean $\pm$ seed s.d., concat-fusion trunk). Bias is train+test; drift and dropout are eval-only. Bold: best per column (oracle excluded). $\dagger$/$\ddagger$: better than no gate at Holm-corrected $p{<}0.05$/$p{<}10^{-6}$ (paired per-take Wilcoxon, seed-averaged takes, $n{=}35$; family of 15: 8 at $p{<}0.05$, 5 at $p{<}10^{-6}$; clean and bias-$\sigma{=}10^\circ$ columns exploratory; full table: supp.~S4). The oracle overrides the per-frame gate arm with whole-take hard masks; attention's bias-$\sigma{=}10^\circ$ cell was not run.}
\label{tab:w2}
\small
\resizebox{\linewidth}{!}{\input{tables/w2_matrix.tex}}
\end{table}

\looseness=-1
Table~\ref{tab:w2} compares the six gate instantiations of Sec.~\ref{sec:method} against two robustness baselines and an oracle; three findings stand out. \emph{(i) Channel-level gating dominates.} The channel gate is the best arm in every column (held-out predictions: Fig.~\ref{fig:qual}): on clean data it beats the ungated control by $16.8$\,mm (Hodges--Lehmann, 95\%\,CI $[10.8,22.4]$) and the strongest stream-level variant (static weights) by $11.9$\,mm $[6.6,20.4]$; under bias, drift, and dropout the HL margins over the ungated model are $+22.1$ $[14.9,29.7]$, $+16.6$ $[11.2,22.8]$, and $+20.5$ $[15.4,26.2]$\,mm (Holm $p{<}10^{-6}$ each; supp.~S4). At script level the margin is sign-consistent in $7/7$ classes in every fault regime ($p{=}0.016$, the $n{=}7$ floor), growing on the seq split ($79.5$ vs.\ $100.7$\,mm clean, seed s0; supp.~S4). \emph{(ii) Stream-level gates add little over static weights.} Static per-stream weights are the strongest stream-level arm under every fault; the temporal and per-frame learned gates numerically match or trail it, and cross-stream softmax attention, which like static is trained without the auxiliary objective, helps modestly under bias and dropout (${\sim}5$\,mm, untested) but not under drift. The aux reliability objective is what carries selectivity: removing it yields the worst stream-level arm on clean data (89.8\,mm, three seeds), and the channel-level variant (single seed) matches clean accuracy (71.7 vs.\ 71.1\,mm) but loses fault robustness (bias $\sigma{=}20^\circ$: 86.5 vs.\ 75.0\,mm) because its orientation gates never close (supp.~S4). A per-frame channel-gate variant under-closes the orientation blocks ($0.83$/$0.61$ clean, seed s0) and trails the temporal channel gate in every regime and seed (three-seed means $92.7$ vs.\ $77.8$\,mm under bias $\sigma{=}20^\circ$, HL $+14.0$\,mm $[+10.4,+18.7]$, $p{=}9.8{\times}10^{-9}$): temporal context, not granularity alone, carries the bias signature (supp.~S4). \emph{(iii) Neither oracle masking nor augmentation alone explains the result.} Overriding the per-frame gate with the ground-truth corruption indicator is catastrophic (132.9\,mm at bias $\sigma{=}20^\circ$): the oracle hard-masks both feet for the whole take, a pattern the trunk never saw in training since dropout augmentation masks only bursts, so test-time masking is a distribution shift. An ungated model with the same corruption augmentation collapses on clean data and under eval-only faults (102.2/100.3/108.3 vs.\ 86.6/87.6/94.1\,mm), helping only under train+test bias, where the fault enters fine-tuning (97.5 vs.\ 100.5\,mm; supp.~S4).

Two disclosures accompany the matrix. The fusion architecture itself costs clean-data accuracy: the concat no-gate control (frozen unit gates) sits at 86.6\,mm where the gate-free trunk of Table~\ref{tab:main} reaches 80.3\,mm (gated-sum: 92.4\,mm). The $6.3$\,mm cost is the per-stream stems and wider fusion projection, and the channel gate's 69.4\,mm more than buys it back. The matrix evaluates foot-stream faults; Sec.~\ref{sec:external} extends to head faults.

\subsection{What the gates learn}
\label{sec:interp}
\begin{figure}[t]
\centering
\includegraphics[width=\linewidth]{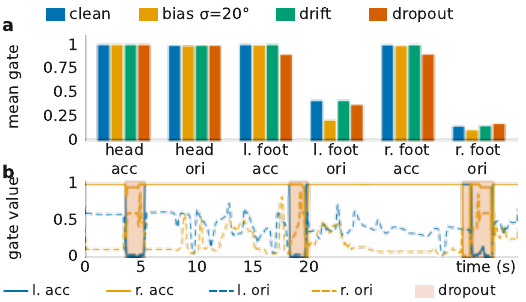}
\caption{Channel gates flag native bias and burst faults without test-time supervision. (a) Mean gate per stream$\times$channel block across regimes: foot orientation gates are partially closed on clean data (native bias) and close further only when bias enters fine-tuning; foot acceleration gates close only under dropout. (b) On a dropout take, foot acceleration gates close inside corrupted bursts (shaded) and reopen.}
\label{fig:gates}
\end{figure}
\looseness=-1
The gates are not supervised at test time, yet they distinguish unreliable channels from reliable ones (Fig.~\ref{fig:gates}; gallery: supp.~S5). On real clean data, with no synthetic corruption at all, the channel gate's foot-orientation blocks sit at $0.41$ (left) and $0.15$ (right) while every other block stays open (${\ge}0.99$). The base selectivity is seeded by the corruption design (bias and drift corrupt only the orientation block; the pretrained gate already partially closes foot orientation on clean synthetic data, $0.66$/$0.65$, supp.~S4), but on real data the closure deepens and lateralizes under the fine-tuning task loss alone (aux off, augmentation active), matching the channel ablation of Sec.~\ref{sec:problem}; a stream-level control without the auxiliary objective leaves gates high and unselective on the same data ($0.81$/$0.93$/$0.91$ per stream): the selectivity is transferred from synthetic-corruption supervision, not emergent. The left/right asymmetry ($0.41{\pm}0.10$ vs.\ $0.15{\pm}0.04$) is unexplained by bias magnitude (median $|B|$ $34.3^\circ$ vs.\ $35.0^\circ$): gate value is channel-specific fault structure, not a calibrated scalar. For constant bias the defense is a learned prior, not online estimation: a fixed model holds accuracy nearly flat across eval-only bias $\sigma{=}5^\circ$--$40^\circ$ ($71.2{\to}77.8$\,mm) with gates nearly unchanged ($0.41{\to}0.44$, seed s0); only bias present in fine-tuning closes them further ($0.21$/$0.11$), and that model keeps the sharpened prior on clean data too ($0.21$/$0.10$): the closure is written into the weights (dose--response: supp.~S4). Under slow drift the stream-level gate fluctuates with local signal statistics without systematic decay (gallery: supp.~S5); the channel gate's closed orientation prior leaves little to track. For burst faults the response is online and fast: foot-acceleration gates close inside dropout bursts (take-averaged $0.08$ vs.\ $1.00$ outside, per-take median drop $-0.94$; Fig.~\ref{fig:gates}b). Zeroed bursts are trivially detectable; the deployment value is one mechanism carrying the native-bias prior and burst response together; the acceleration blocks give $99.5$--$99.7\%$ burst recall at $2$--$6\%$ false positives (AUROC $0.999$/$0.92$; caveats: supp.~S4). A gate-triggered hard fallback is slower to recover and less accurate (supp.~S4): soft down-weighting, not hard switching, is the right use.

\subsection{External baselines and deployment contrasts}
\label{sec:external}
\begin{table}[t]
\centering
\caption{External baselines under eval-only foot corruption (sim-MPJPE mm, run split, bias $\sigma{=}30^\circ$; single seed s0). The channel gate is the most accurate of our learned arms under every fault, at $4.6\times$ HMD-Poser's CPU throughput with $37\%$ fewer parameters (supp.~S7). Bold: best of our learned arms per sim-MPJPE column.}
\label{tab:w3}
\small
\resizebox{\linewidth}{!}{\input{tables/w3_baselines.tex}}
\end{table}

\looseness=-1
Table~\ref{tab:w3} places gated fusion against published systems adapted to our data through a deterministic adapter package (supp.~S3). HMD-Poser~\cite{hmdposer}, a 17.4M-parameter transformer pretrained on full AMASS and fine-tuned per fold under our protocol with its HMD slot fed rotation and rotation velocity only and translational channels zeroed (a handicap; supp.~S3), is the most accurate on clean data (64.4\,mm) but degrades the most under faults ($+16.1$\,mm worst-case vs.\ $+3.5$\,mm for the gate). Paired per-take tests (single seed, exploratory) find HMD-Poser better on clean data (HL $-7.0$\,mm $[-10.7,-2.2]$, $p{=}0.005$) and under drift ($-4.7$ $[-8.4,-0.2]$, $p{=}0.04$; Bonferroni $p{=}0.18$ over the four contrasts), with no significant difference under bias or dropout (HL $+1.8$ $[-6.6,+15.1]$ and $-0.5$ $[-4.0,+3.8]$\,mm; $p{\ge}0.69$). Giving HMD-Poser the same corruption augmentation hurts its clean accuracy (76.2\,mm) and helps no regime: its worst-case degradation stays $+5.9$\,mm (vs.\ $+16.1$\,mm unaugmented and $+3.5$\,mm for the gate). This closes the ``just augment the baseline'' rebuttal. MobilePoser~\cite{mobileposer} zero-shot is insensitive to corruption but inaccurate overall (125.3\,mm); fine-tuning under position-only supervision fails (three recipes, all worse than zero-shot), consistent with its two-stage stack requiring intermediate joint supervision to adapt. Ours trades a slice of clean accuracy for the best worst-case accuracy at $4.6\times$ HMD-Poser's CPU throughput with $37\%$ fewer parameters (supp.~S7), the relevant operating point for always-on deployment, where mounting offsets are large by default and faults accumulate. Two caveats: HMD-Poser's output scale drifts ${\sim}34\times$ under our scale-invariant loss, so the similarity alignment absorbs it and its absolute numbers remain optimistic (supp.~S3); and its clean-data advantage rests on full-AMASS pretraining (vs.\ our three subsets).

\noindent\textbf{A priori drop vs.\ learned gate.}
\looseness=-1
The foot-acceleration-only configuration (feet-A; Sec.~\ref{sec:problem}) is flat under foot faults (66.6--67.6\,mm; orientation faults never reach its inputs, and foot-acceleration dropout costs only $+1.0$\,mm) and nominally beats the channel gate on clean data (66.6 vs.\ 69.4\,mm; the edge persists under multi-view fused labels, supp.~S2), but the comparison crosses trunks. Trunk- and recipe-matched on the gated architecture, the picture inverts: the a priori drop sits flat at $83.4$\,mm and the channel gate beats it on clean, drift, and dropout ($+12.3$/$+12.1$/$+8.6$\,mm, all $p{<}3{\times}10^{-5}$), with no detected difference under strong foot bias ($+4.5$ $[-1.5,+10.9]$, $p{=}0.17$). The residual feet-A advantage is therefore a trunk/recipe effect, not a gating deficit. The decisive contrast is fault location: under head-stream dropout the a priori drop has no mechanism and collapses to 92.9\,mm, while the channel gate partially closes the head-orientation gate inside the bursts ($0.64$ vs.\ $0.99$ outside, seed s0, zero-shot since head faults are never trained) and holds 79.3\,mm (HL $+13.1$\,mm $[+8.3,+18.1]$, $p{=}2{\times}10^{-6}$; matched trunk: $+16.2$\,mm; supp.~S4). Head faults are otherwise milder than foot faults at matched dose (ungated arm, eval-only uniform $\sigma{=}20^\circ$ on both streams: median $+0.9$\,mm head bias vs.\ $+3.8$\,mm foot bias; per-joint breakdown: supp.~S4), consistent with the foot orientation anchoring the distal kinematic chain.

\noindent\textbf{Calibration is per-take, not one-time.}
\looseness=-1
A one-time mounting calibration fitted to each run's opening T-pose segment is worse than no calibration at all (116.7 vs.\ 103.2\,mm; HL $+18.3$\,mm $[+8.2,+26.9]$, $p{=}0.003$), while full per-take calibration reaches 80.3\,mm (supp.~S4): a single static pose does not constrain the sensor-to-bone rotation tightly enough and locks in the residual error. Gating does not remove the need either: without per-take calibration head-only collapses to the mean-pose prior (105.5\,mm) and ungated head+feet nearly does too (103.2\,mm); the channel gate holds 94.0\,mm (seed s0), the best camera-free arm tested but far short of calibrated head-only (79.0\,mm; Table~\ref{tab:main}). Closing that gap without per-take vision is the deployment blocker (Sec.~\ref{sec:discussion}).

\noindent\textbf{Failure modes.}
\label{sec:failure}
\looseness=-1
Per-motion and per-joint breakdowns (supp.~S4, S6) localize the residual error on vertical motion, turning, and hip axial rotation (qualitative predictions: Fig.~\ref{fig:qual}, supp.~S4). Gated fusion bounds the damage from unreliable channels but does not create information: under foot corruption the model falls back on head-dominated behavior, so the residual failure modes are those of the head-only model: geometric (vertical motion and turning; clean head-only $R^2$ $0.06$, carried by the distal joints at $0.49$; the channel gate lifts it to $0.24$) and velocity-blind (every learned model trails the mean-pose prior on MPJVE, 220 vs.\ 253\,mm/s; supp.~S4, S6). A two-logit foot-contact head (BCE against SAM-derived contacts, trained jointly with the pose loss) reaches 0.809 macro-F1 ($+0.02$ over always-contact), and causal unidirectional variants retain selectivity at a uniform accuracy cost (supp.~S4). Drift remains the hardest fault to gate at stream level, since a slow random walk is locally indistinguishable from real motion over a 0.3\,s window; the channel gate's drift margin comes from suppressing the orientation block before drift accumulates, not from detecting drift itself~\cite{sy2021liegroup,hori2025gip}. Out-of-family probes (supp.~S4): the trunk absorbs faults the gates do not detect (gain error, temporal shift, stale bursts), but white accelerometer noise, a fault in the trusted channel, degrades the gated model more than the ungated one ($143$ vs.\ $118$\,mm): the gates keep the noisy acceleration open and mis-attribute the fault. Learned trust allocation concentrates risk in the channels it trusts.

\section{Discussion and Conclusion}
\label{sec:discussion}
\looseness=-1
Consumer sensor reliability is a channel-level property: the same foot stream couples the most valuable signal (acceleration) with the most toxic one (fused orientation), and the standard remedies (augmentation without gating, oracle hard masking, one-time static calibration) all fail. Channel-reliability gating resolves this trade-off without test-time supervision.
\looseness=-1
\emph{Limitations.} All reference signals (alignment, labels, contacts, evaluation) derive from the same monocular model family, so millimeter claims are relative to SAM-3D-Body labels, not true anatomy; headline contrasts are paired on identical labels, so common-mode bias cancels, and residual interactions are partially bounded by the four-view control (Sec.~\ref{sec:data}) and gravity/depth audit (supp.~S1). Temporal alignment is a per-take constant offset with no hardware sync; residual misalignment can mimic orientation-channel toxicity, and within-take drift is unmodeled (excluding low-confidence takes changes no conclusion, supp.~S1). Further: one subject and day, so mechanistic claims plausibly transfer as device properties while millimeter numbers may not; synthetic corruptions; no pelvis sensor (root translation and torso orientation out of reach); per-take vision-derived mounting calibration; and a ${\sim}6$\,mm fusion-architecture cost the channel gate more than recovers. Consumer IMUs fail in ways fixed fusion cannot absorb; learning how much to trust each channel enables deployable capture.

\section*{Acknowledgements}
This work was supported by a UKRI Future Leaders Fellowship [grant number G127262].

{\small
\bibliographystyle{ieee_fullname}
\bibliography{refs}
}

\end{document}

%% file: tables/main_results.tex
\begin{tabular}{lcccccc}
\toprule
Method & sim-MPJPE$\downarrow$ & $R^2_{\text{motion}}\uparrow$ & distal sim$\downarrow$ & distal $R^2\uparrow$ & MPJVE$\downarrow$ & PA-MPJPE$\downarrow$ \\
\midrule
Constant mean-pose prior & 105.5 & 0.03 & 141.2 & -0.07 & 220 & 54.8 \\
Zero-input control (run) & 116.4 & -0.40 & 141.6 & -0.09 & 221 & 63.9 \\
\midrule
\multicolumn{7}{l}{\emph{IMUPoser-adapted}} \\
Synthetic only (zero-shot) & 316.5 & -4.93 & 441.0 & -6.84 & 491 & 218.3 \\
No calibration, head+feet (run) & 103.2 & -0.15 & 124.0 & 0.16 & 317 & 60.6 \\
Head IMU only (run) & 79.0 & 0.06 & 87.0 & 0.49 & 253 & 57.3 \\
Head IMU only (seq) & 92.6 & -0.10 & 107.2 & 0.27 & 249 & 64.3 \\
Head+feet (run) & 80.3 & 0.10 & 90.6 & 0.48 & 287 & 51.6 \\
Head+feet (seq) & 97.6 & -0.10 & 116.4 & 0.18 & 312 & 61.5 \\
\bottomrule
\end{tabular}

%% file: tables/w2_matrix.tex
\begin{tabular}{lccccc}
\toprule
arm & clean & bias $\sigma{=}10^\circ$ & bias $\sigma{=}20^\circ$ & drift & dropout \\
\midrule
no gate (frozen $g{=}1$) & 86.6{\scriptsize$\pm$1.8} & 93.5{\scriptsize$\pm$0.4} & 100.5{\scriptsize$\pm$1.4} & 87.6{\scriptsize$\pm$1.1} & 94.1{\scriptsize$\pm$1.1} \\
stream-dropout training & 88.7{\scriptsize$\pm$0.3} & 92.7{\scriptsize$\pm$0.7} & 98.1{\scriptsize$\pm$2.6} & 89.2{\scriptsize$\pm$0.6} & 92.8{\scriptsize$\pm$1.1} \\
static learned weights & 83.7{\scriptsize$\pm$0.9} & 87.2{\scriptsize$\pm$0.2} & 93.3{\scriptsize$\pm$0.9}$^{\dagger}$ & 83.8{\scriptsize$\pm$1.1} & 85.4{\scriptsize$\pm$0.8}$^{\ddagger}$ \\
cross-stream attention & 87.3{\scriptsize$\pm$0.6} & — & 95.6{\scriptsize$\pm$0.1} & 87.6{\scriptsize$\pm$0.5} & 89.1{\scriptsize$\pm$0.4} \\
per-frame gate, no aux. & 89.8{\scriptsize$\pm$3.7} & 89.5{\scriptsize$\pm$0.8} & 96.5{\scriptsize$\pm$1.2} & 89.5{\scriptsize$\pm$3.8} & 91.4{\scriptsize$\pm$3.4} \\
per-frame gate & 83.9{\scriptsize$\pm$0.7} & 88.9{\scriptsize$\pm$0.4} & 94.3{\scriptsize$\pm$0.7}$^{\dagger}$ & 84.7{\scriptsize$\pm$0.8} & 86.3{\scriptsize$\pm$0.7}$^{\ddagger}$ \\
temporal gate (stream-level) & 86.7{\scriptsize$\pm$0.2} & 94.9{\scriptsize$\pm$3.9} & 97.1{\scriptsize$\pm$0.8} & 87.0{\scriptsize$\pm$0.3} & 88.9{\scriptsize$\pm$0.3}$^{\dagger}$ \\
\textbf{channel gate (ours)} & \textbf{69.4}{\scriptsize$\pm$1.9} & \textbf{74.2}{\scriptsize$\pm$1.7} & \textbf{77.8}{\scriptsize$\pm$1.9}$^{\ddagger}$ & \textbf{69.6}{\scriptsize$\pm$1.8}$^{\ddagger}$ & \textbf{72.8}{\scriptsize$\pm$1.3}$^{\ddagger}$ \\
\midrule
oracle mask override (ref.) & — & — & 132.9{\scriptsize$\pm$0.8} & 141.1{\scriptsize$\pm$6.2} & 97.9{\scriptsize$\pm$1.2} \\
\bottomrule
\end{tabular}

%% file: tables/w3_baselines.tex
\begin{tabular}{lccccc}
\toprule
model & clean & bias & drift & dropout & worst-case \\
\midrule
HMD-Poser (fine-tuned) & 64.4 & 80.4 & 67.8 & 74.2 & +25\% \\
HMD-Poser (ft, corrupt-aug) & 76.2 & 82.1 & 78.8 & 78.1 & +8\% \\
MobilePoser (zero-shot) & 125.3 & 127.6 & 124.7 & 125.2 & +2\% \\
\midrule
no gate & 85.0 & 100.0 & 86.4 & 93.1 & +17.6\% \\
static weights & 83.4 & 87.0 & 83.3 & 85.1 & +4.3\% \\
temporal gate & 86.5 & 90.7 & 86.9 & 88.6 & +4.9\% \\
channel gate (ours) & \textbf{71.1} & \textbf{74.6} & \textbf{71.6} & \textbf{74.3} & +4.9\% \\
\bottomrule
\end{tabular}